\documentclass[sigconf]{acmart}

\AtBeginDocument{%
  \providecommand\BibTeX{{%
    \normalfont B\kern-0.5em{\scshape i\kern-0.25em b}\kern-0.8em\TeX}}}

\copyrightyear{2021}
\acmYear{2021}
\setcopyright{acmcopyright}
\acmConference[ICMR '21]{Proceedings of the 2021
International Conference on Multimedia Retrieval}{August 21--24, 2021}{Taipei,Taiwan}
\acmBooktitle{Proceedings of the 2021 International Conference on Multimedia Retrieval (ICMR '21), August 21--24, 2021, Taipei, Taiwan}
\acmPrice{15.00}
\acmDOI{10.1145/3460426.3463643}
\acmISBN{978-1-4503-8463-6/21/08}

\usepackage{graphicx}
\usepackage{float}
\usepackage{caption,subcaption}
\usepackage{amsmath}

\usepackage{amssymb}
\newcommand{\cmark}{\checkmark}
\newcommand{\xmark}{$\times$}

\begin{document}

\fancyhead{}
\title{Few-Shot Action Localization without Knowing Boundaries}


\author{Ting-Ting Xie, Christos~Tzelepis}
\email{t.xie@qmul.ac.uk, c.tzelepis@qmul.ac.uk}
\affiliation{
  \institution{Queen Mary University of London}
  \streetaddress{Mile End Road}
  \city{London}
  \country{UK}}


\author{Fan Fu}
\email{fan.fu@city.ac.uk}
\affiliation{
      \institution{City, University of London}
      \streetaddress{Northampton Square}
      \city{London}
      \country{UK}}

\author{Ioannis Patras}
\email{i.patras@qmul.ac.uk}
\affiliation{
  \institution{Queen Mary University of London}
  \streetaddress{Mile End Road}
  \city{London}
  \country{UK}}

\renewcommand{\shortauthors}{Ting-Ting Xie, et al.}


\begin{abstract}
Learning to localize actions in long, cluttered, and untrimmed videos is a hard task, that in the literature has typically been addressed assuming the availability of large amounts of annotated training samples for each class -- either in a fully-supervised setting, where action boundaries are known, or in a weakly-supervised setting, where only class labels are known for each video. In this paper, we go a step further and show that it is possible to learn to localize actions in untrimmed videos when a) only one/few trimmed examples of the target action are available at test time, and b) when a large collection of videos with only class label annotation (some trimmed and some weakly annotated untrimmed ones) are available for training; with no overlap between the classes used during training and testing. To do so, we propose a network that learns to estimate Temporal Similarity Matrices (TSMs) that model a fine-grained similarity pattern between pairs of videos (trimmed or untrimmed), and uses them to generate Temporal Class Activation Maps (TCAMs) for seen or unseen classes. The TCAMs serve as temporal attention mechanisms to extract video-level representations of untrimmed videos, and to temporally localize actions at test time. To the best of our knowledge, we are the first to propose a weakly-supervised, one/few-shot action localization network that can be trained in an end-to-end fashion. Experimental results on THUMOS14 and ActivityNet1.2 datasets, show that our method achieves performance comparable or better to state-of-the-art fully-supervised, few-shot learning methods.

\end{abstract}

\ccsdesc[500]{Computing methodologies~Machine learning}

\keywords{action localization; few-shot learning; weakly-supervised learning}

\maketitle

\section{Introduction}\label{sec:introduction}

Localizing actions in videos is a challenging task that has received increasing attention in the last years~\cite{oikonomopoulos2009implicit,shou2016temporal,gao2017turn,gao2018ctap,lin2019bmn,xu2020g,paul2018wtalc,shou2018autoloc,liu2019completeness,shi2020weakly,yang2018one,feng2018video}. A central challenge in this field, is the difficulty in obtaining large scale, fully annotated data, where the temporal extend of the different actions are given as ground truth. To address this issue, several recent works have appeared on topics such as weakly supervised localization~\cite{wang2017untrimmednets,paul2018wtalc,shou2018autoloc,narayan20193c,liu2019completeness,shi2020weakly}, few-shot action detection~\cite{yang2018one,xu2020revisiting} and video re-localization~\cite{feng2018video,huang2020weakly,yang2020localizing}.


\begin{figure*}[t!]
    \centering
    \includegraphics[width=0.9\textwidth]{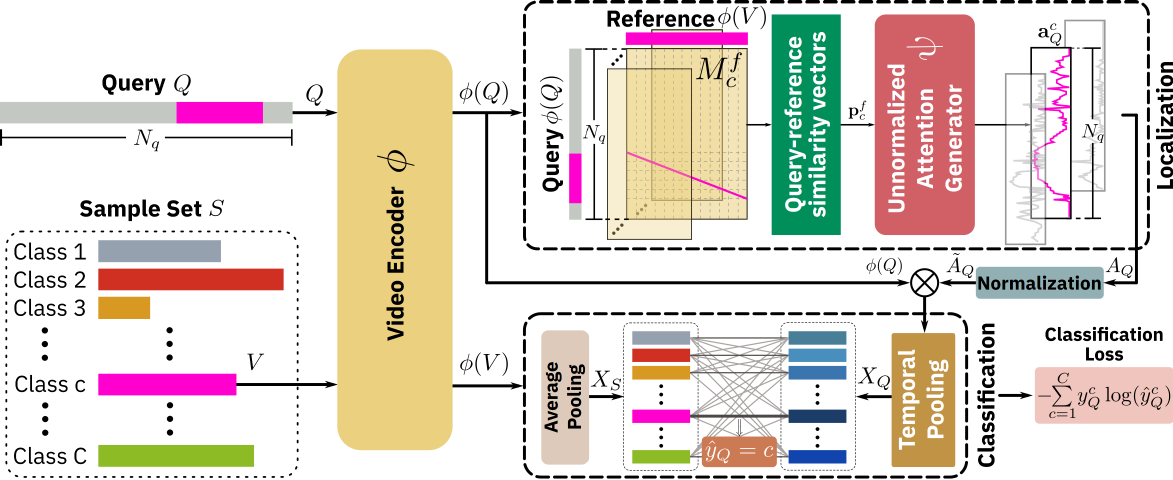}
    \vspace{-2mm}
    \caption{Overview of the proposed method ($C$-way 1-shot): Given a query video $Q$ and a sample set (of trimmed reference videos), we refine their feature representations using an encoder $\phi$ and learn attention masks (TCAMs) by learning generator $\psi$ on a set of query-reference similarity vectors calculated by Temporal Similarity Matrices (TSMs). We perform localization by keeping those consecutive snippets of the query with attention values greater than a certain threshold and classify the localized actions by comparing to reference video features.}
    \label{fig:overview}
\end{figure*}

Few-shot learning~\cite{fei2006one} has been used in several domains, including action recognition. Such methods~\cite{bishay2019tarn,Cao2020few,zhang2020few}, typically rely on learning a similarity function between pairs of videos on a training set and use it to compare videos in the test set with videos in a support set that contains one or few examples of novel classes (i.e., classes that have not been seen during training). In the domain of video action localization, the few recent few-shot learning approaches that have been published (e.g.,~\cite{yang2018one,xu2020revisiting}), do so by assuming fully annotated training examples, i.e., known temporal borders of the classes on the query set during training. 

In these works, this information is used to train a class-agnostic first stage proposal generator and/or as a supervision signal to the similarity function that is learned between pairs of snippets in the support and the query videos and/or to select the snippets in the untrimmed query videos on which the similarity is learned. However, manual annotation of the borders of actions is time-consuming and sometimes ambiguous.

To address the problem that temporal annotation of action borders is a time-consuming task, several weakly-supervised learning methods~\cite{paul2018wtalc,narayan20193c,shi2020weakly} have been proposed. These methods split the video into snippets (e.g., 16 frames) and perform classification at snippet level to obtain temporal class activation maps (TCAMs)~\cite{zhou2016learning}. Those maps are used during training as attention mechanisms to refine the classifiers, and during testing to localize the actions. However, such methods~\cite{paul2018wtalc,narayan20193c,shi2020weakly} rely on classifiers that are learned for the classes that are present in the training set each of which has typically several samples. This is very different from the one/few-shot learning framework, where only one/few samples are available for the classes in the test set; in such cases, training a classifier is impractical/prone to over-fitting.

To address these problems, we propose a weakly supervised method for one/few-shot action localization, that is a method that localizes actions in untrimmed videos when a) only one/few trimmed examples of the target action are available at test time, and b) when a large collection of videos with class label (some trimmed and some weakly annotated untrimmed) are available for training -- clearly, without overlap between the classes used during training and testing. We do so by designing a network that during training learns a similarity function that estimates Temporal Similarity Matrices (TSMs), that is, fine-grained snippet-to-snippet level similarity patterns between pairs of videos (trimmed or untrimmed). These are subsequently used in order to  generate Temporal Class Activation Maps (TCAMs) for seen or unseen classes. The TCAMs serve as temporal attention mechanisms to extract video-level representations of untrimmed videos at training time, and to temporally localize actions in them at test time. 

Our TCAMs are similar in functionality with those in other weakly supervised works ~\cite{nguyen2018weakly,nguyen2019weakly,shi2020weakly}, however, a crucial difference is that, in our case, TCAMs are calculated based on similarities with reference videos as in~\cite{kordopatis2019visil}, and not from class-based classifiers that are hardly trained from one/few examples. During training, we optimize a classification loss at video level, in order to ensure the inter-class separability of learned features. This is in contrast to other works on few-shot action localisation that at training, they have fine-grained action labels at snippet level and therefore can supervise their similarity function at the level of action-proposals~\cite{yang2018one,xu2020revisiting}, whose overlap with the ground truth is known. We show that with the proposed method we obtain similar, or better performance than them, even though they are trained in a fully-supervised manner, i.e., with the annotation of action boundaries in the untrimmed videos in the training set.

Our main contributions are summarized as follows: Firstly, we address a novel and challenging task, namely \textit{weakly-supervised few-shot video action localization}, which attempts to locate instances of unseen actions using one/few examples by learning from videos (trimmed and untrimmed), with only video-level labels. To the best of our knowledge, we are the first to address this problem. Secondly, by contrast to other weakly supervised methods, we propose an end-to-end single stage method to generate Temporal Class Activation Maps (TCAMs) from \textit{Temporal Similarity Matrices (TSMs)} and not from class-based classifiers. This allows the generation of TCAMs using sample-query video pairs from both seen and unseen classes, and avoids additional proposal generation stage. Last but not the least, our results are comparable or better to those of fully-supervised few-shot action localization methods.


\section{Related Work}\label{sec:related_work}

Traditional fully-supervised deep learning methods typically require large amounts of annotated data, introducing a significant prone-to-ambiguity annotation workload~\cite{zhao2019hacs,xie2020temporal,shao2020finegym,shao2020intra}. For this reason, learning with scarce data (i.e., few-shot learning) has received increasing attention, in domains like object detection~\cite{fei2006one,vinyals2016matching,sung2018learning,sun2019meta,hou2019cross,michaelis2020closing}, action recognition~\cite{zhu2018towards,hahn2019action2vec,bishay2019tarn,Cao2020few,Brattoli2020rethinking,zhang2020few}, and action localization~\cite{yang2018one,feng2018video,huang2020weakly,yang2020localizing}. Current works in this domain either learn using trimmed~\cite{kordopatis2017near,zhu2018compound,bishay2019tarn,Cao2020few,zou2020compositional,Brattoli2020rethinking} or well-annotated untrimmed videos~\cite{yang2018one}, or address class-agnostic localization tasks~\cite{feng2018video,huang2020weakly,yang2020localizing} -- learning with both scarce data and limited annotation for both action recognition and localization is still an under-explored area. 

\subsection{Temporal action localization}\label{subsec:rel_work_tal}

Video action localization has been extensively studied under the fully-supervised paradigm~\cite{shou2016temporal,zhao2017temporal,gao2017turn,lin2018bsn,long2019gaussian,lin2019bmn,xu2020g}. However, due to the challenging, time-consuming, and prone-to-ambiguity task of data collection and annotation, weakly-supervised approaches have received increasing attention by the research community~\cite{wang2017untrimmednets,shou2018autoloc,paul2018wtalc,nguyen2018weakly,narayan20193c,shi2020weakly,min2020adversarial}. In this case, video annotation is given only with respect to the video-level action class, while the exact boundaries of the class instances are not available during training. 

More specifically, driven by the effectiveness of fully-supervised two-stage temporal action localization methods~\cite{zhao2017temporal,gao2017cascaded,lin2018bsn,lin2019bmn}, recent works~\cite{wang2017untrimmednets,shou2018autoloc} propose to classify a set of candidate proposals by training a video-level classifier. For instance, UntrimmedNet~\cite{wang2017untrimmednets} generates proposals by uniform or shot-based sampling that are subsequently fed to a classification module trained on video-level labels. AutoLoc~\cite{shou2018autoloc} generates temporal class activation maps (TCAMs) by performing video-level classification and arrives at TCAM-based proposals using an appropriate loss function during training the localization model. 

In contrast to the above, some works have directed efforts towards improving TCAM for improving weakly-supervised temporal action localization. For instance,~\cite{paul2018wtalc,narayan20193c} propose to exploit the correlations between similar actions, and~\cite{nguyen2018weakly} imposes background suppression. \cite{min2020adversarial} proposes the optimization of a two-branch network in an adversarial manner so as one branch localizes the most salient activities of a video, while the other discovers supplementary ones, from non-localized parts of the video. \cite{shi2020weakly} propose to discriminate the action and context frames by a conditional VAE~\cite{kingma2013vae} by maximizing the likelihood of each frame with respect to the attention values.

\subsection{Few-shot learning}\label{subsec:rel_work_fsl}

Few-shot learning paradigm has been extensively studied for video related tasks, such as action recognition~\cite{zhu2018compound,bishay2019tarn,Cao2020few,zou2020compositional,Brattoli2020rethinking}. CMN~\cite{zhu2018compound} utilizes the key-value memory network paradigm to obtain an optimal video representation in a large space, then classifies videos by matching and ranking. TARN~\cite{bishay2019tarn} and OTAM~\cite{Cao2020few} exploit the temporal information missed from previous few-shot learning methods~\cite{zhu2018compound,careaga2019metric} by imposing temporal alignment before measuring distances. Zou et al.~\cite{zou2020compositional} propose a soft composition mechanism to investigate compositional recognition that human can perform, which has been well studied in cognitive science, but not well explored under few-shot learning setting. Brattoli et al.~\cite{Brattoli2020rethinking} conduct an in-depth analysis of end-to-end training and pre-trained backbones for zero-shot learning.

Recently, few-shot learning has been adopted also for the problem of video action localization~\cite{yang2018one,xu2020revisiting} under the fully-supervised paradigm. \cite{yang2018one,xu2020revisiting} uses a two-stage approach where, in the first stage, it applies a proposal generator to generate class-agnostic action proposals, and in the second stage it feeds them to a network that learns to compare them (using some similarity metric) to the categorical samples for classification. The difference between~\cite{yang2018one,xu2020revisiting} and us is, the supervision signal in~\cite{yang2018one,xu2020revisiting} is much stronger. Knowing the overlap of each proposal with ground truth segments during training, they are able to distinguish actions from background explicitly in loss function. By contrast, we adopt a weakly-supervised setting, extract class-specific video-level representations of the untrimmed videos using the TCAMs as attention masks, and learn using a video-level classification cost. Besides, \cite{yang2018one,xu2020revisiting} exploit a proposal generation stage, learning or not, while we do not.



In a recently proposed line of research, video re-localization, Feng et al.~\cite{feng2018video} propose to localize in a query video segments that correspond semantically to a given reference video. Huang et al.~\cite{huang2020weakly} extends the original formulation so as to learn without using temporal boundaries information in the training set by utilizing a multi-scale attention module. Besides, Yang et al.~\cite{yang2020localizing} assume only one class in each query video and more than one support videos. Different with~\cite{feng2018video,huang2020weakly,yang2020localizing}, we work on multi-class video localization, focusing on not only localization but also classification, which is more challenging than single-class example-based re-localization -- assuming only one action, from the same class with reference video, to be located in a given query video.


\section{Few shot, weakly supervised localization}\label{sec:method}

In this paper, we address the problem of weakly-supervised few-shot action localization in videos. In this framework, the training set contains videos that are annotated with only class label(s), both trimmed and untrimmed ones, possibly more than one labels per video, and typically contains a large number of examples of each class. During testing, we are given a support set that contains one/few examples of novel classes, and a test set that contains untrimmed videos in which we seek to localize the actions of those novel classes. Adopting the protocol followed by~\cite{vinyals2016matching,sung2018learning,yang2018one}, we consider $C$-way $K$-shot episode training/testing. More specifically, in each episode, we randomly select $C$ classes from the training set and $K$ trimmed action instances for each class to serve as \textbf{sample set} $\mathcal{S}$, and untrimmed videos with video-level annotations as \textbf{query set} $\mathcal{Q}$, in each at least one action instance of the $C$ classes exists. During a test episode, given a query video, the task is to generate snippet-level attention masks, and categorize each snippet into one of the $C$ classes or as background.

\subsection{Proposed method}\label{subsec:proposed_method}

Our method consists of two learnable modules, namely the \textit{video encoder} $\phi$ and the \textit{attention generator} $\psi$. The video encoder is used to generate meaningful embeddings in order to calculate a small set of temporal similarity matrices (TSMs) between query (from query set) and reference videos (from sample set) using different similarity metrics, which subsequently are used in order to learn attention masks. An overview of the proposed method is given in Fig.~\ref{fig:overview}. 

Given a query video $Q=\left(\mathbf{q}_1,\ldots,\mathbf{q}_{N_q}\right)\in\mathbb{R}^{n\times N_q}$ and a reference video $V=\left(\mathbf{v}_1,\ldots,\mathbf{v}_{N_v}\right)\in\mathbb{R}^{n\times N_v}$ from the sample set, where $\mathbf{q}_i,\mathbf{v}_i\in\mathbb{R}^{n\times1}$ denote their $i$-th snippet, respectively, represented by using either the RGB or the optical flow features~\cite{carreira2017quo}. Note that, $n$ is the feature dimension and $N_*$ is the number of snippets of the corresponding video. We first use the video encoder $\phi$ in order to transform them into embeddings $\phi(Q)=\left(\phi(\mathbf{q}_1),\ldots,\phi(\mathbf{q}_{N_q})\right)\in\mathbb{R}^{d\times N_q}$ and $\phi(V)=\left(\phi(\mathbf{v}_1),\ldots,\phi(\mathbf{v}_{N_v})\right)\in\mathbb{R}^{d\times N_v}$, respectively. Note that we train a separate video encoder for each feature representation scheme (RGB and optical flow).

Subsequently, we obtain the TSMs by calculating pair-wise embedding similarities for all snippet pairs between the query and reference video, using various similarity metrics (i.e., we compute one TSM for each similarity metric choice and each class). With a max-pooling operation along the time dimension of the reference video in TSM we obtain the similarity of each of the $N_q$ snippets of the query video with the reference video. We arrive at the attention masks by learning the attention generator $\psi$ module that takes as input four similarity vectors, one for each combination of features (RGB and optical flow) and similarity metrics (dot product and cosine distance). By setting a threshold on these attention masks we assign action/background labels to each snippet; this way, we localize actions at snippet-level (localization block in Fig.~\ref{fig:overview}).

For doing classification, we compare the transformed (using video encoder $\phi$) reference videos, after applying a pooling operation in order to fix their dimensions, to the product of the normalized attention masks and transformed (using the same video encoder $\phi$) query features, in order to decide on the class of the action that are previously localized (classification block in Fig.~\ref{fig:overview}). Below, we will discuss each part of the proposed method in detail.

\subsection{Video Encoder}\label{subsec:video_encoder}

As described above, the video encoder $\phi$ is used in order to refine pre-trained features and arrive at representations more meaningful to the task at hand. More specifically, we use I3D~\cite{carreira2017quo} as a pre-trained feature extractor, similarly to~\cite{nguyen2018weakly,paul2018wtalc}. I3D incorporates both spatial and temporal information by using two stream of RGB and TV-L1 optical flow~\cite{zach2007tvl1} -- this has been shown to benefit activity detection~\cite{xie2019exploring,chao2018rethinking}. We give non overlapping two-stream $16$-frame snippets as input and pass its output through a 3D pooling layer of kernel size $2\times7\times7$ in order to obtain $1024$-dimensional features in each stream. The video encoder $\phi$ consists of two Fully-Connected (FC) layers, with output dimensions $1024$ and $128$ respectively, each layer using a ReLU activation function. To avoid over-fitting, we use dropout after the first FC layer.

\subsection{Temporal similarity and attention generation}\label{subsec:tsm_attention}

\begin{figure}[t]
    \centering
    \includegraphics[width=0.45\textwidth]{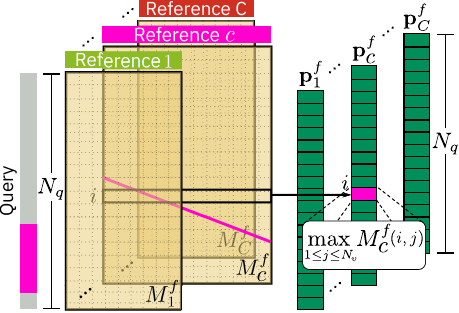}
    \vspace{-2mm}
    \caption{The Temporal Similarity Matrix (TSM) $M^f_c$ is calculated on the snippets of a pair of the query and a reference video from class $c\in\{1,\ldots,C\}$ using a similarity metric $f$. Then, a single score is assigned to each snippet $i$ of the query video by max-pooling along the $i$-th row of $M^f_c$, i.e., $\mathbf{p}^f_c(i)=\max_{1\leq j\leq N_v}M^{f}_c(i,j)$.}
    \label{fig:tsm}
\end{figure}

Given the embeddings $\phi(Q)\in\mathbb{R}^{d\times N_q}$ of the given query and $\phi(V)\in\mathbb{R}^{d\times N_v}$ of a reference video of class $c\in\{1,\ldots,C\}$, we calculate a snippet-to-snippet similarity matrix $M_c^f\in\mathbb{R}^{N_q\times N_v}$, which we call Temporal Similarity Matrix (TSM). More specifically, the $(i,j)$-th entry of $M_c^f$, i.e., the similarity between the snippets $\phi(\mathbf{q}_i)$ of the query and $\phi(\mathbf{v}_j)$ of the reference video, is given as $M^f_c(i,j)=f(\phi(\mathbf{q}_i),\phi(\mathbf{v}_j))$, where $f$ is a similarity metric. Given $M_c^f\in\mathbb{R}^{N_q\times N_v}$, we then assign a single similarity score $\mathbf{p}^f_c(i)$ to each snippet $i$ of the query video that expresses how well the snippet matches the reference video. We do so, by max-pooling along the rows of $M^f_c$ (see Fig.~\ref{fig:tsm}), that is, 
\begin{equation}\label{eq:sim_value}
    \mathbf{p}^f_c(i)=\max_{1\leq j\leq N_v}M^{f}_c(i,j), \quad i=1,\ldots,N_q.
\end{equation}

\begin{figure}[t]
\centering
    \includegraphics[width=0.4\textwidth]{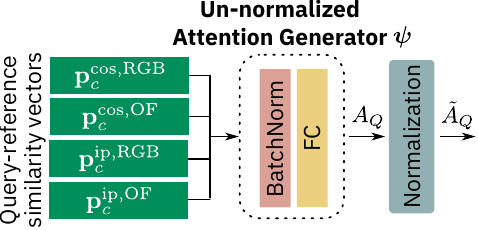}
    \vspace{-2mm}
    \caption{The attention generator $\psi$ consists of a pair of batch normalization and FC trainable layers and takes as input the concatenation of the cosine and dot product similarities computed on both RGB and optical flow features. The attentions ($A_Q$) are used for localization while normalized ones ($\tilde{A}_Q$) are used for classification, as described in Sect.~\ref{subsec:localization_classification}.}
    \label{fig:attention_generator}
\end{figure}

In practice, we calculate four TSMs for each class: one for each combination of two distances (cosine and dot product) and two types of features (RGB and optical flow). By doing so, we arrive at four similarity vectors $\mathbf{p}^{\cos,\text{RGB}}_{c}$, $\mathbf{p}^{\cos,\text{OF}}_{c}$, $\mathbf{p}^{\text{ip},\text{RGB}}_{c}$, and $\mathbf{p}^{\text{ip},\text{OF}}_{c}$ using (\ref{eq:sim_value}) for each class $c\in\{1,\ldots,C\}$. We found that beneficial in comparison to using only a single type of similarity metric and/or feature. The similarities are then concatenated and given as input to an attention generator module consisting of a batch normalization and a FC layer (Fig.~\ref{fig:attention_generator}). The output Temporal Class Attention Mask (TCAM) is then given by
\begin{equation}\label{eq:attention_mask}
    \mathbf{a}_Q^c=\psi(\mathbf{p}^{\cos,\text{RGB}}_{c},\mathbf{p}^{\cos,\text{OF}}_{c},\mathbf{p}^{\text{ip},\text{RGB}}_{c},\mathbf{p}^{\text{ip},\text{OF}}_{c})\in\mathbb{R}^{N_q\times1}.
\end{equation}
Finally, by normalizing each $\mathbf{a}_Q^c$ to $\tilde{\mathbf{a}}_Q^c$ using the softmax operator, we arrive at normalized Temporal Class Attention Masks, as
\begin{equation}
    \tilde{A}_Q=\left(\tilde{\mathbf{a}}_Q^1,\ldots,\tilde{\mathbf{a}}_Q^C\right)\in\mathbb{R}^{N_q\times C}.
\end{equation}

\subsection{Localization and Classification}\label{subsec:localization_classification}

Training and testing of our method is done in $C$-way, $K$-shot episodes where, at each episode, an untrimmed query video $Q$ is compared to $C \times K$ videos in the sample set -- the latter set contains $K$ examples of $C$ randomly sampled classes. In this section, we will show how we obtain action localisation maps and scores, and how we obtain video level scores for the query video. To simplify the notation, we will first present the 1-shot scenario and then how it can be trivially extended to the $K$-shot case. 

\textbf{Localization} After obtaining the TCAM $\mathbf{a}_Q^c$ for a query video $Q$ and class $c$, we threshold the $\mathbf{a}_Q^c$ and group together consecutive snippets that are above a given threshold $\delta$. Then, following the standard practice~\cite{paul2018wtalc,narayan20193c,shi2020weakly}, we arrive at a set of action predictions $(s, e, p)$, where $s, e$ and $p$ are the start, end, and prediction score of a certain prediction. We set the prediction score as the average of $\mathbf{a}_Q^c$ of the individual snippets, that is, $p=\frac{1}{e-s+1}\sum_{i=s}^{e+1} \mathbf{a}_Q^c(i)$. In the case of $K$-shot, for a specific class $c$, we average $K$ TCAMs calculated from $K$ samples to be the final TCAM $\mathbf{a}_Q^c$. We use $\mathbf{a}_Q^c$ here for the sake of its high discriminative ability among snippets compared with $\tilde{\mathbf{a}}_Q^c$ (normalised TCAM).

\textbf{Classification} We use the normalized TCAMs $\tilde{\mathbf{a}}_Q^c$ in order to obtain class specific vector representations for the query video. More specifically, for class $c$, we multiply (element-wise) $\tilde{\mathbf{a}}_Q^c$ (by broadcasting $\tilde{\mathbf{a}}_Q^c$ along the $N_q$-dimensional vector (\ref{eq:attention_mask}) as an $N_q\times d$ matrix) with the transformed features $\phi(Q)$, leading to a $N_q\times d$ matrix. By summing up over video length (i.e., weighted temporal average pooling~\cite{nguyen2018weakly}) for each class, we arrive at a set of $C$ $d$-dimensional vectors, each of which corresponds to the class-wise representation of the query video. This is $X_Q\in\mathbb{R}^{C\times d}$, depicted by the blue toned vectors in the classification block of Fig.~\ref{fig:overview}.

At the same time, we transform each of the videos in the sample set using the video encoder $\phi$ and apply a temporal average pooling operation in order to fix their dimensionality (note that videos from sample set are typically of different lengths) to $d$. Thus, we arrive at a representation $X_S\in\mathbb{R}^{C\times d}$ for the reference videos $S$ in the sample set (see Fig.~\ref{fig:overview}). The final score of the query $Q$ for class $c$ is then given by
\begin{equation}
    \hat{y}_Q^c = \frac{\exp\left(-\lVert X_Q(c,:)-X_S(c,:)\rVert^2\right)}{\sum_{j=1}^{C}\exp\left(-\lVert X_Q(j,:)-X_S(j,:)\rVert^2\right)},
\end{equation}
where $X_Q(j,:)$, $X_S(j,:)$ denote the $j$-th rows of $X_Q$ and $X_S$, respectively. In $K$-shot, we calculate $K$ distances for each class, which we average and proceed as above.

\paragraph{Classification loss} Since we adopt a weakly-supervised setting, we only use video-level class labels. Given that $y_Q^c$ denotes the ground truth label of query video with respect to class $c$ and $\hat{y}_Q^c$ its predicted label given as described above, we optimize a cross-entropy loss term given as follows
\begin{equation}\label{eq:classification_loss}
    \mathcal{L}_{\text{cls}} = -\sum_{c=1}^C y_Q^c\log(\hat{y}_Q^c).
\end{equation}

\section{Experiments}\label{sec:experiments}

\textbf{Datasets} We evaluate the proposed method on two popular datasets for video action localization, namely THUMOS14~\cite{THUMOS14} and ActivityNet1.2~\cite{caba2015activitynet}. \textbf{THUMOS14} provides annotations for 101 classes and consists of 1010 validation (THUMOS14-Val) and 1574 testing videos (THUMOS14-Test). However, temporal annotations (for 20 classes) are provided only for 200 validation and 213 testing videos. This is typically referred to as the THUMOS14-Val-20/Test-20 split. Following the standard practice~\cite{zhao2017temporal,yang2018one,shi2020weakly}, we train our models on the validation set and evaluate them on the testing set for the boundary localization task. \textbf{ActivityNet1.2} provides annotations (both in terms of video-level class and temporal boundaries) for 100 classes on 4819 training (ANET-Train) and 2383 validation videos (ANET-Val). Following the standard protocol (e.g.,~\cite{wang2017untrimmednets,shou2017cdc,xu2017r,shi2020weakly}), we train our models on the training set and evaluate on validation set. Note that we do not use any temporal boundaries information during training, but only during testing for evaluating our models. Similarly to~\cite{yang2018one}, we use trimmed videos in the support set, but in contrast to them we do not utilize the temporal boundary annotations in the query videos.

\textbf{Few-shot training/evaluation protocol} Few-shot learning paradigm requires that the classes used for testing must not be present during training. Following~\cite{yang2018one}, for THUMOS14, we we use a part of THUMOS14 validation set (6 classes from THUMOS14-Val-20) to train both video encoder $\phi$ and attention generator $\psi$ (see Fig.~\ref{fig:overview}). We use the remaining 14 classes from the test set (THUMOS14-Test-14) in order to evaluate our one-shot localization network. For ActivityNet1.2, we split the 100 classes into 80/20 splits. We train our localization network on 80 classes in the training set, denoted as ANET-Train-80, and  evaluate on the other 20 classes in the validation set, denoted as ANET-Test-20, following~\cite{yang2018one}.

\textbf{Training} In each training episode, we randomly choose 5 classes from the set of training classes, and train our network under the standard $5$-way $K$-shot setting~\cite{sung2018learning} using 8 query videos for each class. In THUMOS14 ($5$-way $5$-shot), we use 5 query videos due to limited amount of data. In each training episode ($C$-way $K$-shot), our model will be trained in a mini-dataset of $C$ classes, in which we split into two non-overlap subsets. One of them consists of $K$ videos from each class, and we use one trimmed action instance from each to form the sample set; the rest untrimmed videos will be served as query videos. 

\textbf{Testing} Our (meta-)testing setting is similar to that of (meta-)training, except for the support and testing sets. More specifically, we pair a randomly chosen video in testing set with 5 examples. Due to the large number of different combinations of 5 examples (random classes/samples from each class), and since the localization performance relies on them, similarly to~\cite{yang2018one}, we randomly sample 1000 different examples from each of the test classes and calculate mAP across all these examples. In experiments, we report the median of 10 repetitions.


\textbf{Evaluation metrics} Following the literature in temporal action localization, we evaluate our models using mean Average Precision (mAP) at different temporal Intersection over Union (tIoU) thresholds (mAP@tIoU). In ActivityNet1.2, we also report the average mAP at 10 evenly distributed tIoU thresholds between 0.5 and 0.95~\cite{zhao2017temporal,shou2018autoloc,shi2020weakly}. We also report the numerical video-level action recognition accuracy of top-1 and top-3 predicted classes. 

\textbf{Implementation details} We train our network using the Adam optimizer~\cite{kingma2014adam} with an initial learning rate of $10^{-4}$, which we decrease by a factor of 2 after 1000 episodes, a weight-decay factor of $5\cdot10^{-4}$, and a dropout rate of $0.5$. For both datasets, we train for 10000 episodes.

\subsection{Main results}

\begin{table*}[]
    \centering
    \caption{Temporal action localization performance of the proposed method in terms of mAP (\%) for various tIoU thresholds, 5-way, one-shot (@1), and 5-shot (@5).}
    \label{tab:tious}
    \begin{tabular}{cccccccccccccc} \hline
        tIoU & 0.1 & 0.2 & 0.3 & 0.4 & 0.5 & 0.6 & 0.7 & 0.8 & 0.9 & Top-1 & Top-3\\ \hline
        THUMOS14@1 & 31.31 & 27.48 & 23.19 & 18.64 & 13.93 & 9.78 & \textbf{6.55} & \textbf{3.02} & \textbf{0.61} & 52.15 & 86.65\\ 
        THUMOS14@5 & \textbf{34.28} & \textbf{31.19} & \textbf{25.98} & \textbf{19.70} & \textbf{14.20} & \textbf{9.91} & 6.06 & 2.87 & 0.46 & \textbf{59.30} & \textbf{91.50} \\ \hline
        ActivityNet1.2@1 & 64.57 & 59.18 & 55.14 & 51.16 & 45.76 & 40.85 & 35.01 & 28.03 & 17.95 & 74.50 & 97.65 \\ 
        ActivityNet1.2@5 & \textbf{73.21} & \textbf{67.17} & \textbf{62.80} & \textbf{57.90} & \textbf{52.59} & \textbf{46.18} & \textbf{38.75} & \textbf{31.61} &	\textbf{19.82} & \textbf{83.50} & \textbf{99.20}\\ \hline
    \end{tabular}
\end{table*}

\begin{table}[]
    \centering
    \caption{Comparison of the proposed method with fully-supervised few-shot learning state-of-the-art methods on THUMOS14 in terms of mAP@0.5, one-shot (@1), and five-shot (@5).}
    \label{tab:soa_fsl_th14}
    \begin{tabular}{ccc} \hline
        Supervision & Method & mAP@0.5    \\ \hline
        Full & CDC@1~\cite{shou2017cdc} & 6.4 \\ 
        Full & CDC@5~\cite{shou2017cdc} & 6.5 \\ 
        Full & Sl. window@1~\cite{yang2018one} & 13.6 \\
        Full & Sl. window@5~\cite{yang2018one} & 14.0 \\ 
        Full &  F-PAD@1~\cite{xu2020revisiting} & 24.8\\ 
        Full &  F-PAD@5~\cite{xu2020revisiting} & 28.1\\ \hline
        Weak & Ours@1 & 13.9 \\
        Weak & Ours@5 & 14.2 \\ \hline
    \end{tabular}
\end{table}
\begin{table}[t]
    \centering
    \caption{Comparison of the proposed method with fully-supervised few-shot learning state-of-the-art methods on ActivityNet1.2 in terms of mAP@0.5, average of mAP@0.5:0.95 (avg), one-shot (@1), and five-shot (@5).}
    \label{tab:soa_fsl_anet}
    \begin{tabular}{cccc} \hline
        Supervision & Method                                & mAP@0.5  & avg  \\ \hline
        Full        & CDC@1~\cite{shou2017cdc}              & 8.2      & 2.4  \\ 
        Full        & CDC@5~\cite{shou2017cdc}              & 8.6      & 2.5  \\ 
        Full        & Sl. window@1~\cite{yang2018one}       & 22.3     & 9.8  \\
        Full        & Sl. window@5~\cite{yang2018one}       & 23.1     & 10.0 \\ 
        Full        &  F-PAD@1~\cite{xu2020revisiting}        & 41.5     & 28.5 \\ 
        Full        &  F-PAD@5~\cite{xu2020revisiting}        & 50.8     & 34.2 \\ \hline
        Weak        & Ours@1                                & 45.8     & 31.4 \\
        Weak        & Ours@5                                & 52.6     & 35.3 \\ \hline
        
    \end{tabular}
\end{table}
We evaluate our method on THUMOS14 and ActivityNet1.2 and compare with state-of-the-art \textit{fully-supervised} few-shot methods for the lack of other \textit{weakly-supervised} few-shot methods. We report results on THUMOS14 and ActivityNet1.2 datasets in Tables~\ref{tab:soa_fsl_th14} and~\ref{tab:soa_fsl_anet}, respectively. More specifically, on THUMOS14, we surpass~\cite{shou2017cdc} by a large margin for both 1-shot and 5-shot 5-way settings, while we achieve very similar results with~\cite{yang2018one}. Besides, table~\ref{tab:soa_fsl_th14} also shows our model lags behind F-PAD~\cite{xu2020revisiting}, which most likely due to the proposal generation subset they trained on the boundary information. On ActivityNet1.2, we outperform both fully-supervised works~\cite{shou2017cdc,yang2018one}, by a large margin -- particularly we outperform the state-of-the-art~\cite{yang2018one,xu2020revisiting} in both $1$- and $5$-shot 5-way settings (e.g., in the case of $5$-shot, we achieve a mAP@0.5 of 52.6\% compared to 23.1\% of~\cite{yang2018one} and 50.8\% of~\cite{xu2020revisiting}).  

It is worth noting that the different performance in two datasets is due to their different relative difficulty in the context of temporal action localization~\cite{shi2020weakly}. That is, THUMOS14 consists of more fine-grained action instances per video (15.5 on average), compared to ActivityNet1.2 (1.5 on average). Moreover, action instances in THUMOS14 typically range from a few seconds to minutes, making them, in practice, sparsely distributed in a clutter of backgrounds, compared to ActivityNet1.2 where the actions are long and typically of only one class in each video. This is also reflected in Table~\ref{tab:tious}, where we report localization results in terms of mAP for different tIoU thresholds. We see that we achieve much higher localization performance for small tIoUs in ActivityNet1.2 compared to THUMOS14. This is also observed in top-1 classification accuracy.

The above have informed our choice of the threshold $\delta$ that we set on the TCAMs. For THUMOS14, we use the middle of the range $\frac{\max(A_Q^c)+\min(A_Q^c)}{2}$. For the ActivityNet1.2, we use different thresholds for the different classes in the sample set. More specifically, we set the threshold for the class $c$ such that the average length of the predictions in the query video is similar to the (average) length of the action of that class in the support video. Finally, same to~\cite{shi2020weakly}, we note THUMOS14 has fewer weakly annotated videos for training.

\subsection{Ablation studies}

We conduct a number of ablation studies in order to demonstrate the effectiveness of a) the two main learnable components of our method, namely the video encoder $\phi$ and the attention generator $\psi$, and b) various secondary design choices, such as the similarity metric. We choose to evaluate on THUMOS14, since it is more challenging than ActivityNet1.2, under the 5-way, 1-shot setting.

\begin{figure*}[t!]
    \centering
    \includegraphics[width=\textwidth]{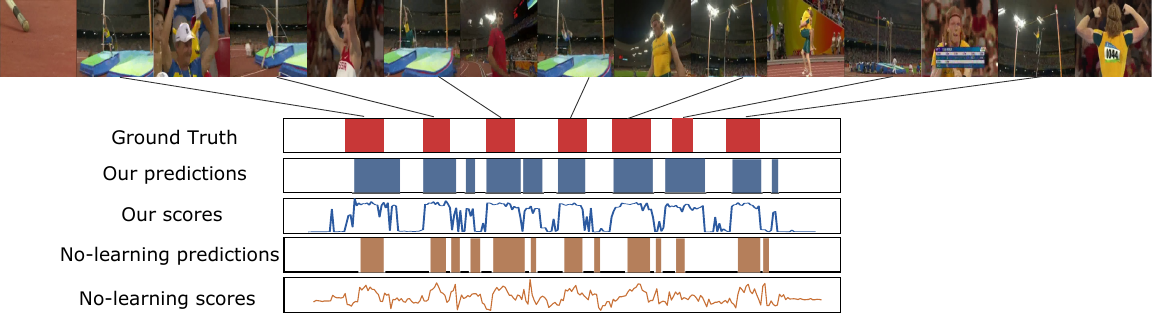}
    \vspace{-2mm}
    \caption{Examples of predicted proposals and TCAMs $A_Q$, learnt and not learnt, for ground truth action \textit{Pole Vault}. Rows in blue are when the video encoder $\phi$ and the attention generator $\psi$ are optimized during training, and in brown are when pre-trained I3D~\cite{carreira2017quo} features are directly used for calculating the temporal similarity matrices (TSMs) and attention masks (TCAMs). (*It is \textit{video\_test\_0000444} from THUMOS14.)}
    \label{fig:attention_masks}
\end{figure*}

\begin{table}[]
    \centering
    \caption{THUMOS14 5-way, $K$-shot evaluation result without learning video encoder $\phi$ and attention generator $\psi$.}
    \label{tab:no_learn}
    \begin{tabular}{p{0.1cm}p{1.4cm}p{1.2cm}ccc} \hline
        $K$ & Similarity  & Pooling   & mAP@0.5           & Top-1              & Top-3         \\ \hline
        1 & Euclidean     & weighted  & 2.11              & 27.78             & 70.40          \\
        1 & Cosine        & weighted  & 6.50              & 41.14             & 76.73          \\ 
        1 & Dot Prod.     & average   & 10.24             & 41.29             & 76.72          \\ 
        1 & Dot Prod.     & weighted  & \textbf{10.24}    & \textbf{47.75}    & \textbf{85.20} \\ \hline
        5 & Dot Prod.     & weighted  & \textbf{11.83}    & \textbf{53.95}    & \textbf{87.70} \\ \hline
    \end{tabular}
\end{table}

\textbf{Without learning} We begin by evaluating our architecture without learning the video encoder $\phi$ or the attention generator $\psi$ (see Fig.~\ref{fig:overview}). More specifically, we do this by using directly the pre-trained I3D~\cite{carreira2017quo} in order to calculate the temporal similarity matrices (TSMs) and attention masks (TCAMs), as described in Sect.~\ref{sec:method}. In Table~\ref{tab:no_learn} we report the performance of our network, in terms of action recognition accuracy and localization mAP@0.5, when no learning is conducted, for various combinations of similarity metrics and temporal pooling operations. We note that using the dot product or the cosine distance for calculating TSMs, outperforms Euclidean distance by large margins with respect to both classification and localization. Moreover, in order to investigate the effectiveness of weighted temporal average pooling (in order to calculate the video-level representations as in Fig.~\ref{fig:overview}), we compare it with average pooling. We see that using TCAMs improves by 6.46\% and 8.48\% the top-1 and top-3 action recognition accuracy.



\begin{table}[]
    \centering
    \caption{Ablation study of video encoder ($\phi$) and attention generator ($\psi$) on THUMOS14 (5-way, 1-shot).}
    \label{tab:abl_arch}
    \begin{tabular}{cccccc} \hline
        $\phi$  & $\psi$ & Similarity & mAP@0.5        & Top-1           & Top-3          \\ \hline
        \xmark  & \xmark & Dot Prod.  & 10.24          & 47.75           & 85.20          \\ 
        \cmark  & \xmark & Dot Prod.  & 9.60          & \textbf{51.80}  & 84.55          \\
        \xmark  & \cmark & -          & 11.45         & 47.65           & 81.85          \\
        \cmark  & \cmark & -          & \textbf{13.93} & 51.70           & \textbf{86.65} \\ \hline
    \end{tabular}
\end{table}

\begin{table}[]
    \centering
    \caption{Ablation studies on different $C$ and $K$ ($C$-way $K$-shot) on THUMOS14.}
    \label{tab:abl_ck}
    \begin{tabular}{ccccc} \hline
        $C$ & $K$ & mAP@0.5 & Top-1 & Top-3 \\ \hline \hline
        1 & 1 & 27.37 & - & - \\ 
        1 & 5 & 27.83 & - & - \\ 
        1 & 10 & \textbf{27.90} & - & - \\ \hline
        10 & 1 & 10.42 & 39.50 & 69.50 \\
        10 & 5 & 12.99 & 52.65 & 78.45\\
        10 & 10 & \textbf{13.85} & \textbf{52.75} & \textbf{80.85}\\ \hline
        14 & 1 & 9.09 & 34.55 & 59.35\\ 
        14 & 5 & 10.80 & 45.70 & 69.85\\ 
        14 & 10 & \textbf{11.76} & \textbf{49.40} & \textbf{74.00}\\ \hline
    \end{tabular}
\end{table}

\textbf{Learning $\phi$ and $\psi$} Next, we proceed into investigating the effectiveness of training the video encoder $\phi$ and the attention generator $\psi$ modules. In Table~\ref{tab:abl_arch}, we report the localization performance, in terms of mAP@0.5, and the classification performance, in terms of the top-1 and top-3 accuracy, on THUMOS14 under the 5-way, 1-shot setting. We note that training the video encoder alone improves the action recognition ability of our network (e.g., top-1 accuracy is improved from 47.75\% to 51.80\%). Moreover, training the attention generator alone (Fig.~\ref{fig:attention_generator}) improves the localization performance by 1.21\%. Finally, training both the video encoder and the attention generator arrives at better performance both in terms of localization (+3.69\%) and recognition accuracy (top-1: +3.95\%, top-3: +1.45\%).

\textbf{$C$-way, $K$-shot} To investigate the generalization ability of our method, we test with different $C$ and $K$ parameters ($C$-way, $K$-shot) using the model we trained using the $5$-way, $1$-shot setting (Table~\ref{tab:abl_ck}). Compared to $C=5$ (Table~\ref{tab:tious}), as expected, localization performance of $C=1$ is increased, since under this setting the problem boils down to class-agnostic action localization. In the cases of $C=10$ or $C=14$, even though classification task is more challenging, which leads to an anticipated drop in classification performance, we note that our method achieves slightly worse or comparable localization performance. 

\textbf{Visualization} We conclude our ablation studies by illustrating how training the learnable modules, i.e., video encoder and attention generator, affects the attention masks used for temporal action localization. In Fig.~\ref{fig:attention_masks} we show an indicative example of the attention masks (multiplied by the video class scores) to which our method arrives when we learn the video encoder and attention generator (in blue), and the corresponding masks when we do not learn any of them (in brown). We note that when we optimize $\phi$ and $\psi$, we arrive at more meaningful attention masks, which subsequently lead to better segmentation of the query video with respect to ground truth. It is also worth noting that background snippets are suppressed in the case of learnt attention masks. 

\subsection{1-shot results on novel splits.}
\label{subsec:disc}

\begin{figure*}[ht]
    \centering
    \begin{subfigure}[b]{0.45\linewidth}
        \includegraphics[height=5cm,width=\textwidth]{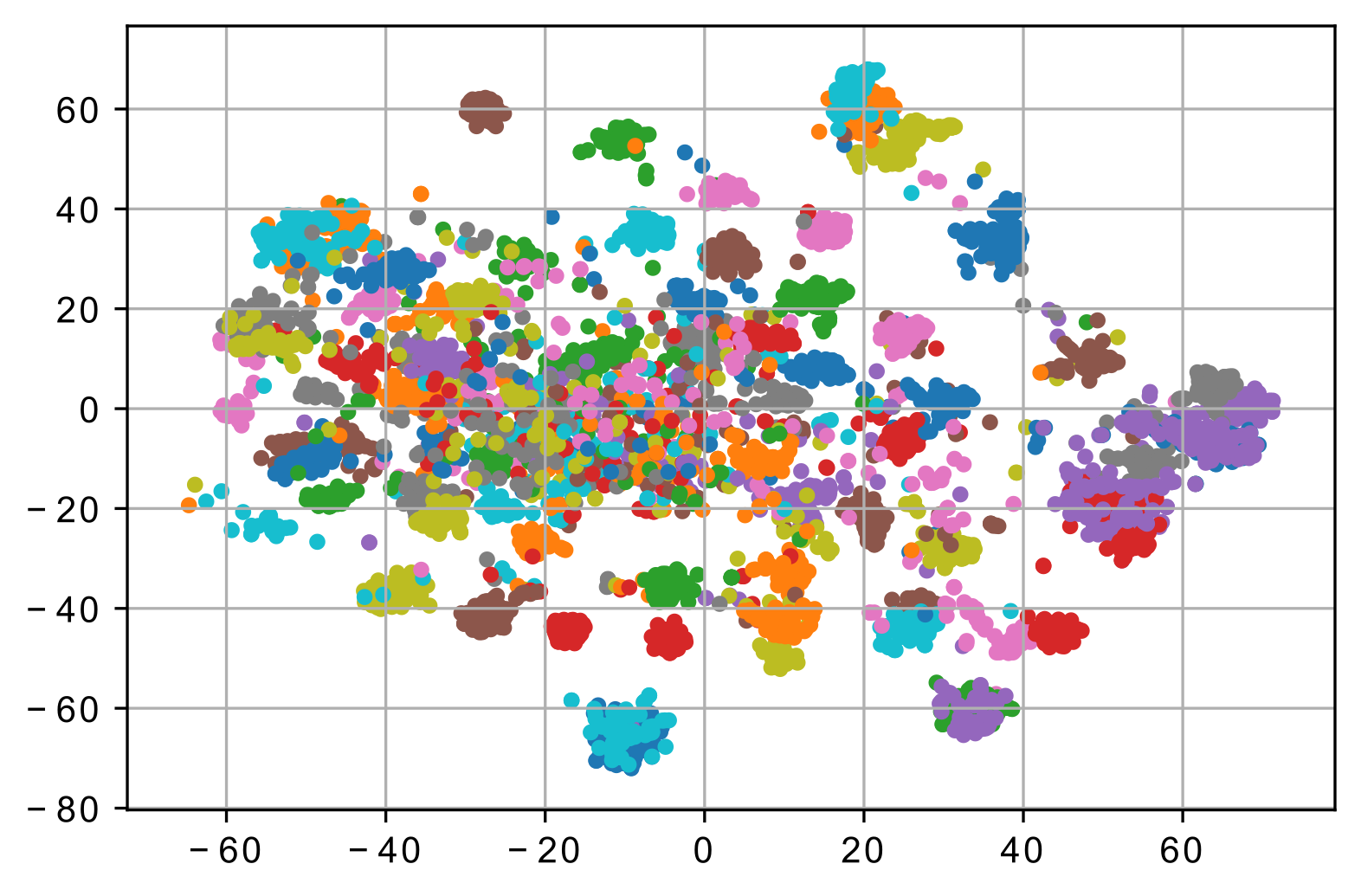}
        \caption{}
        \label{subfig:atrain}
    \end{subfigure}
    \begin{subfigure}[b]{0.45\linewidth}
        \includegraphics[height=5cm,width=\textwidth]{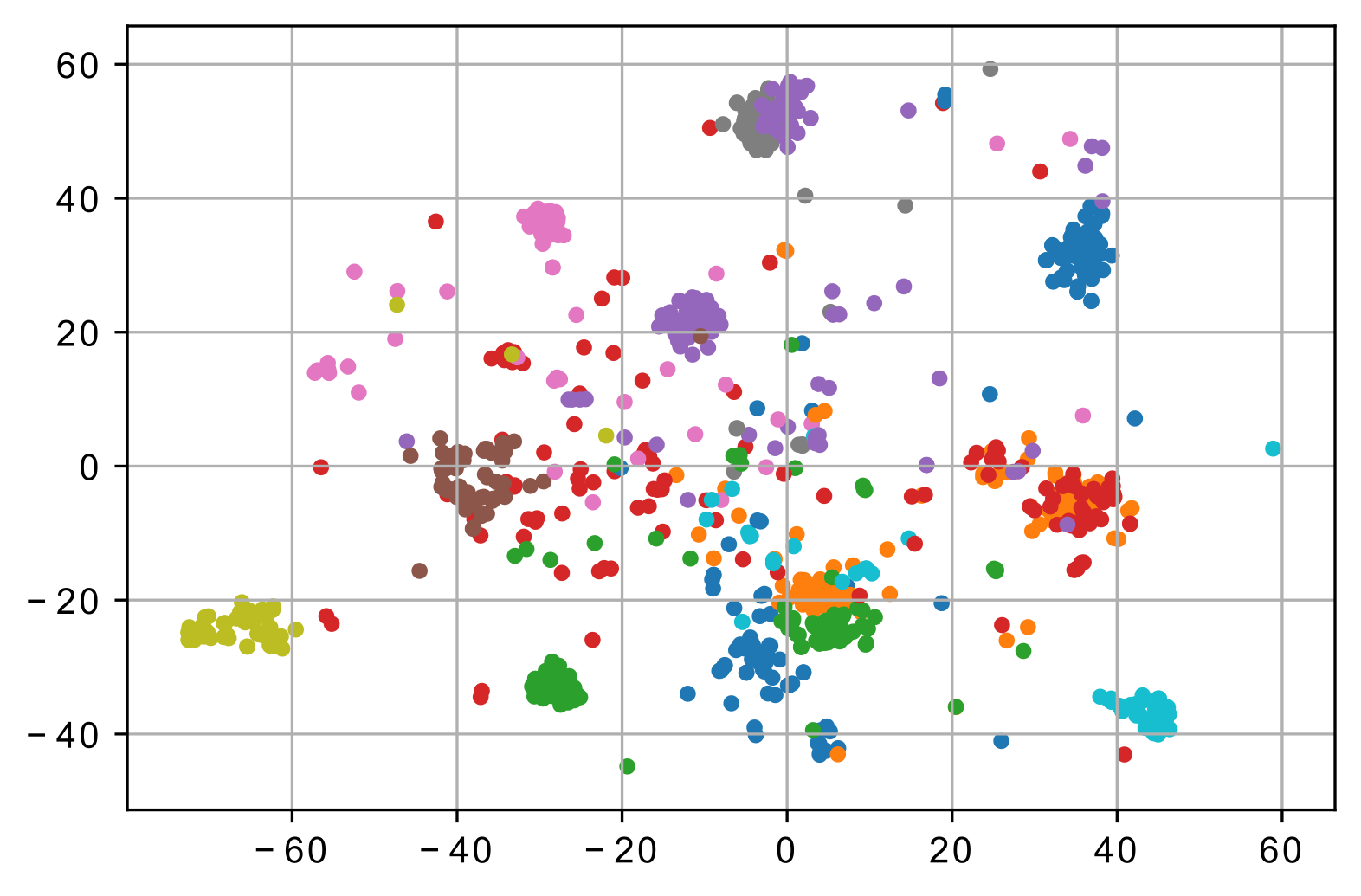}
        \caption{}
        \label{subfig:atest}
    \end{subfigure} 

\vspace{-2mm}
\caption{Feature visualization on ActivityNet1.2. left: In-Domain classes. right: Out-of-Domain classes. Each point represents a video with label from In-Domain (or Out-of-Domain) class set, while the feature is calculated by: 1) Applying average pooling operation across temporal dimension; 2) Reducing feature dimension by t-SNE.}
\label{fig:anet_tsne}
\end{figure*}

\begin{figure}[htp]
    \centering
    \includegraphics[height=4.5cm,width=\linewidth]{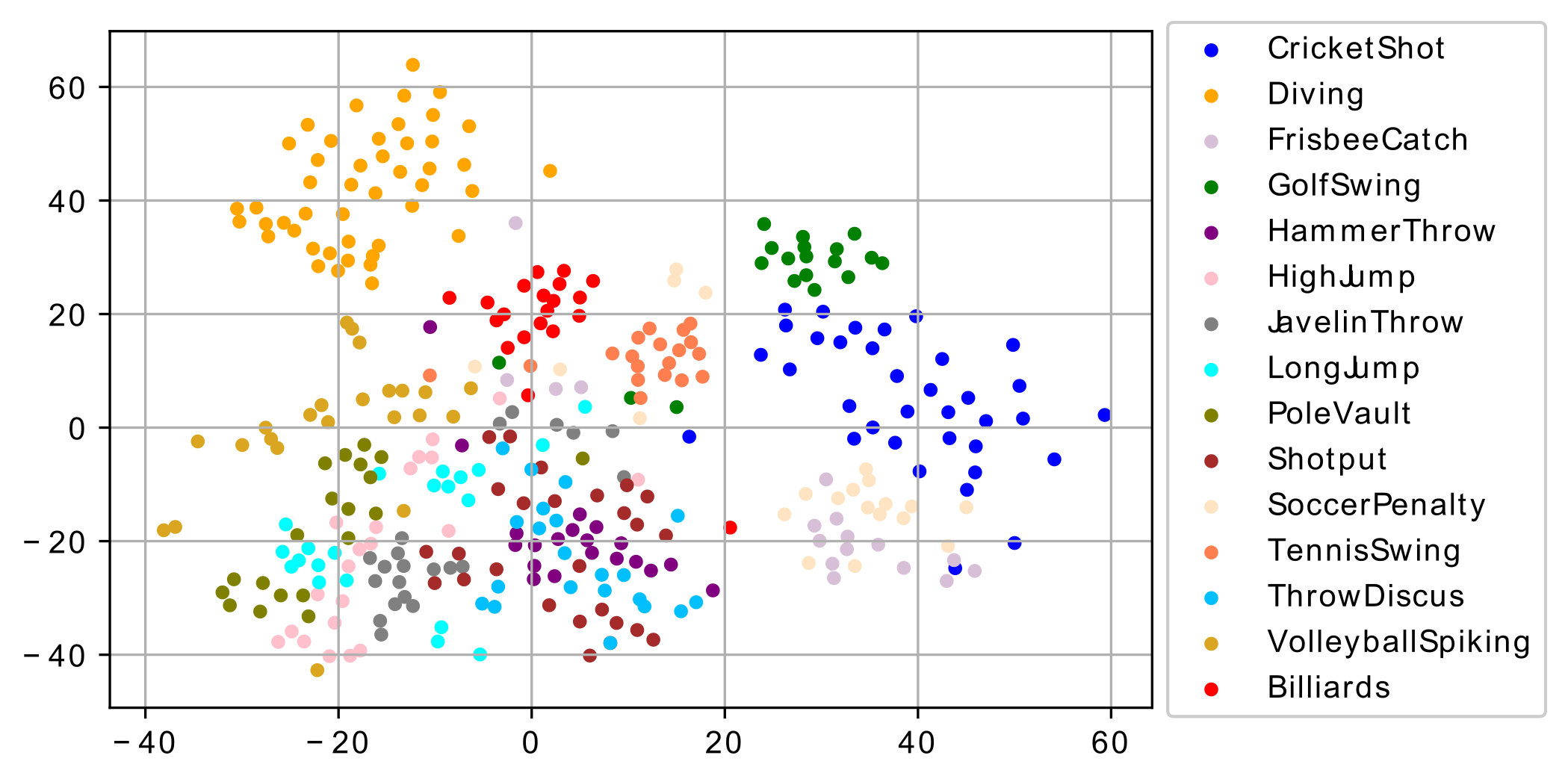}
    \vspace{-2mm}
    \caption{Out-of-Domain t-SNE feature visualization on THUMOS14.}
    \label{fig:th14_tsne}
\end{figure}

Usually few-shot learning refers to the study of generalizing to unseen categories in image/action classification. This paper, following~\cite{larochelle2008zero,radford2learning}, instead uses ``\verb|few-shot learning|" in a broader sense and mainly studies generalization to unseen datasets/tasks, which is closer to ``\verb|few-shot transfer|". More precisely, in the experiments reported in the previous section, we utilized the pre-trained features on Kinetics-400~\cite{kay2017kinetics} from~\cite{carreira2017quo}, a dataset which has overlapped actions with THUMOS14 and ActivityNet1.2. 

To further demonstrate the effectiveness of our method, this section reports results on a ``novel split" that uses In-Domain classes (overlapped with kinetics-400) as training set and Out-of-Domain classes (non-overlapped with kinetics-400) as testing set. We show that with the help of the representations trained in a large dataset, our model yields comparable performance on both In-/Out-of-Domain classes in other datasets.  

For Table~\ref{tab:AnetTALonunseen}, we train on 28 In-Domain classes and evaluate on 72 Out-of-Domain classes in ActivityNet1.2. The results show similar recognition ability as in Table~\ref{tab:soa_fsl_anet} and an expected drop over localization mAP due to the decrease in the size of the training set. Still, the result outperforms~\cite{yang2018one} and \cite{xu2020revisiting} by a large margin (39.48\% vs 22.3\%\cite{yang2018one} vs 31.7\%\cite{xu2020revisiting}). Fig.~\ref{fig:anet_tsne} shows Kinetics-400 pretrained feature on ActivityNet1.2, indicating that the feature itself is useful to discriminate actions, no matter whether they belong to the in- or out-of-domain subsets of Kinetics-400. From the THUMOS14-20 classes, only ``Billiards" is not present in Kinetics-400, so we remove 'Billiards' from the training set and add it to evaluation set -- we see that there is only a marginal drop in the performance of this class (Table~\ref{tab:TH14TALonunseen}). As seen in Fig.~\ref{fig:th14_tsne} where the T-SNE plots of classes is presented, even though ``Billiards" is not present in Kinetics-400, it is still distinguishable, which is consistent with the visualization of the ActivityNet1.2 classes. Taking analysis above and results without learning (Table~\ref{tab:no_learn}) into consideration, we believe that a good feature representation (such as one obtained by pretraining on Kinetics) is the key to localization performance with few examples.

\begin{table}[]
    \centering
    \caption{Results(\%) of 5-way 1-shot on ActivityNet1.2. (Note, origin split is the same as~\cite{yang2018one}; F-PAD~\cite{xu2020revisiting} is pretrained on a larger dataset -- sports-1M~\cite{karpathy2014large}, * is the controlled split testing on Out-of-Domain classes, which is different from `origin'.)}
    \begin{tabular}{cccccc} \hline
         split  & \#train/test      & mAP@0.5 & avg   & Top-1 & Top-3 \\ \hline
         F-PAD~\cite{xu2020revisiting}*  & 80/20 & 31.7 & 19.4 & - & - \\ 
         origin & 80/20         & 45.76   & 31.43 & 74.50 & 97.65 \\
         novel & 72/28         &39.48   & 25.21 & 74.20 & 96.00 \\ \hline
    \end{tabular}
    \label{tab:AnetTALonunseen}
\end{table}

\begin{table}[]
    \centering
    \caption{Results(\%) of 5-way 1-shot on THUMOS14. (Note, \cmark means 'Billiard' is used for training; \xmark for testing. \# indicates number of classes for training/testing.)}
    \begin{tabular}{ccccccc} \hline
         Billiards & \#train/test  & mAP@0.5 & Top-1 & Top-3 \\ \hline
         \cmark    &     6/14      & 13.93   & 52.15 & 86.65 \\
         \xmark    &     5/15      & 12.97   & 51.83 & 86.12 \\ \hline
    \end{tabular}
    \label{tab:TH14TALonunseen}
\end{table}

\section{Conclusion}\label{sec:conclusion}

In this paper, we proposed a weakly-supervised few-shot method for the problem of temporal action localization in videos. To the best of our knowledge, this is the first method to address this problem under the assumptions of few-shot learning using  video-level annotation only. We do that by learning to estimate meaningful temporal similarity matrices that model fine grained similarity patterns between pairs of videos (trimmed or untrimmed), and use them to generate attention masks for seen or unseen classes. Experimental results on two datasets show that we our method achieves performance comparable to SoA fully-supervised, few-shot learning methods.

\begin{acks}
Funding for this research is provided by the joint QMUL-CSC Scholarship, and by the EPSRC Programme Grant EP/R025290/1.

\end{acks}

\bibliographystyle{ACM-Reference-Format}
\bibliography{icmr}

\end{document}